\documentclass[runningheads]{llncs}
\usepackage{graphicx}
\usepackage{amsmath,amssymb} 
\usepackage{color}
\usepackage[width=122mm,left=12mm,paperwidth=146mm,height=193mm,top=12mm,paperheight=217mm]{geometry}
\usepackage{times}
\usepackage{epsfig}
\usepackage{graphicx}
\usepackage{amssymb}
\usepackage{graphicx}
\usepackage{setspace}

\usepackage{multirow}
\usepackage{booktabs}

\usepackage{calrsfs}
\DeclareMathAlphabet{\pazocal}{OMS}{zplm}{m}{n}

\newcommand{\Ib}{\pazocal{I}}	
\newcommand{\Vb}{\pazocal{V}}

\newcommand{\R}{\mathbb{R}}

\begin{document}
\pagestyle{headings}
\mainmatter

\title{Spatio-Temporal Channel Correlation Networks \\ for Action Classification} 

\titlerunning{A very long title}

\authorrunning{authors running}

\author{Ali Diba$^{1,4,\star}$, Mohsen Fayyaz$^{3,\star}$, Vivek Sharma$^{2}$, M.Mahdi Arzani$^{4}$, Rahman Yousefzadeh$^{4}$, Juergen Gall$^{3}$, Luc Van Gool$^{1,4}$}


\institute{$^{1}$ESAT-PSI, KU Leuven, $^{2}$CV:HCI, KIT, Karlsruhe, $^{3}$University of Bonn, $^{4}$Sensifai \\ 
$^{1}$\{firstname.lastname\}@kuleuven.be, $^{2}$\{firstname.lastname\}@kit.edu, $^{3}$\{lastname\}@iai.uni-bonn.de, $^{4}$\{firstname.lastname\}@sensifai.com}

\maketitle

\begin{abstract}
The work in this paper is driven by the question if spatio-temporal correlations are enough for 3D convolutional neural networks (CNN)? Most of the traditional 3D networks use local spatio-temporal features. We introduce a new block that models correlations between channels of a 3D CNN with respect to temporal and spatial features. This new block can be added as a residual unit to different parts of 3D CNNs. We name our novel block 'Spatio-Temporal Channel Correlation' (STC). By embedding this block to the current state-of-the-art architectures such as ResNext and ResNet, we improved the performance by 2-3\% on Kinetics dataset. Our experiments show that adding STC blocks to current state-of-the-art architectures outperforms the state-of-the-art methods on the HMDB51, UCF101 and Kinetics datasets.
The other issue in training 3D CNNs is about training them from scratch with a huge labeled dataset to get a reasonable performance. So the knowledge learned in 2D CNNs is completely ignored. Another contribution in this work is a simple and effective technique to transfer knowledge from a pre-trained 2D CNN to a randomly initialized 3D CNN for a stable weight initialization. This allows us to significantly reduce the number of training samples for 3D CNNs. Thus, by fine-tuning this network, we beat the performance of generic and recent methods in 3D CNNs, which were trained on large video datasets, e.g. Sports-1M, and fine-tuned on the target datasets, e.g. HMDB51/UCF101. 
\footnote{$^{\star}$Ali Diba and Mohsen Fayyaz contributed equally to this work. Mohsen Fayyaz contributed to this work while he was at Sensifai.}
\end{abstract}

\section{Introduction} \label{sec:intro}
Compelling advantages of exploiting temporal rather than merely spatial cues for video classification have been shown lately~\cite{tle,c3d,n3d}. In recent works, researchers have focused on improving modeling of spatio-temporal correlations. Like 2D CNNs, 3D CNNs try to learn local correlation along input channels. Therefore, 3D CNNs neglect the hidden information in between channels correlations in both directions: space and time, which limits the performance of these architectures. Another major problem in using 3D CNNs is training the video architectures calls for extra large labeled datasets. All of these issues negatively influence their computational cost and performance. To avoid these limitations, we propose (i) a new network architecture block that efficiently captures both spatial-channels and temporal-channels correlation information throughout network layers; and (ii) an effective supervision transfer that bridges the knowledge transfer between different architectures, such that training the networks from scratch is no longer needed.

Motivated by the above observations, we introduce the spatio-temporal channel correlation (STC) block. The aim of this block is considering the information of inter channels correlations over the spatial and temporal features simultaneously. For any set of transformation in the network (e.g. convolutional layers) a STC block can be used for performing spatio-temporal channel correlation feature learning.The STC block has two branches: a spatial correlation branch (SCB) and a temporal correlation branch (TCB). The SCB considers spatial channel-wise information while TCB considers the temporal channel-wise information. The input features $I \in \mathbb{R}^{H\times W \times T \times C}$ are fed to SCB and TCB. In SCB a spatial global pooling operation is done to generate a representation of the global receptive field which plays two vital roles in network: (i) considering global correlations in $I$ by aggregating the global features over the input, (ii) providing a channel-wise descriptor for analyzing the between channels correlations. This channel-wise feature vector is then fed to two bottleneck fully connected layers which learn the dependencies between channels. The same procedure happens in TCB, however, for the first step a temporal global pooling is used instead of the spatial global pooling. Output features of these two branches are then combined and returned as the output of the STC block. These output features can be combined with the output features of the corresponding layer(s).  By employing such features along-side traditional features available inside a 3D CNN, we enrich the representation capability of 3D CNNs. Therefore, the STC block equipped 3D CNNs are capable of learning channel wise dependencies which enables them to learn better representations of videos. We have added the STC block to the current state-of-the-art 3D CNN architectures such as 3D-ResNext and 3D-ResNet \cite{3DResHara}. The STC block is inserted after each residual block of these networks.
 
As mentioned before, training 3D CNNs from scratch need a large labeled dataset. It has been shown that training 3D Convolution Networks ~\cite{c3d} from scratch takes two months~\cite{res3d} for them to learn a good feature representation from a large scale dataset like Sports-1M, which is then finetuned on target datasets to improve performance. Another major contribution of our work therefore is to achieve supervision transfer across architectures, thus avoiding the need to train 3D CNNs from scratch. Specifically, we show that a 2D CNN pre-trained on ImageNet can act as `\textit{a teacher}' for supervision transfer to a randomly initialized 3D CNN for a stable weight initialization. In this way we avoid the excessive computational workload and training time. Through this transfer learning, we outperform the performance of generic 3D CNNs (C3D~\cite{c3d}) which was trained on Sports-1M and finetuned on the target datasets HMDB51 and UCF101.

The rest of the paper is organized as follows. In Section~\ref{sec:related}, we discuss related work. Section~\ref{sec:method} describes our proposed approaches. The implementation details,  experimental results and their analysis are presented in Section~\ref{sec:experiments}. Finally, conclusions are drawn in Section~\ref{sec:conclusion}.

\section{Related Work} \label{sec:related} 

\noindent
\textbf{Video Classification with and without CNNs:}
Video classification and understanding has been studied for decades. Several techniques have been proposed to come up with efficient spatio-temporal feature representations that capture the appearance and motion propagation across frames in videos, such as HOG3D~\cite{ref15}, SIFT3D~\cite{ref23}, HOF~\cite{ref18}, ESURF~\cite{ref38}, MBH~\cite{ref4}, iDTs~\cite{ref34}, and more. These were all hand-engineered. Among these, iDTs yielded the best performance, at the expense of being computationally expensive and lacking scalability to capture semantic concepts. It is noteworthy that recently several other techniques~\cite{rankbasura} have been proposed that also try to model the temporal structure in an efficient way. 

Using deep learning, the community went beyond hand-engineered representations and learned the spatio-temporal representations in an end-to-end manner. These methods operate on 2D (frame-level) or 3D (video-level) information. In the 2D setting, CNN-based features of individual frames are modeled via LSTMs/RNNs to capture long-term temporal dependencies~\cite{lstm1,n3d}, or via feature aggregation and encoding using Bilinear models~\cite{tle}, VLAD~\cite{actionvlad}, Fisher encoding~\cite{fishernet} etc. 
Recently, several temporal architectures have been proposed for video classification, where the input to the network consists of either RGB video clips or stacked optical-flow frames. The filters and pooling kernels for these architectures are 3D (x, y, time). The most intuitive are 3D convolutions~($s \times s\times d$)~\cite{n3d} where the kernel temporal depth $d$ corresponds to the number of frames used as input, and $s$ is the kernel spatial size. Simonyan et al.~\cite{twostream} proposed a two-stream network, cohorts of RGB and flow CNNs. In their flow stream CNNs, the 3D convolution has $d$ set to 10. Tran et al.~\cite{c3d} explored 3D CNNs with filter kernel of size $3\times 3 \times 3$. Tran et al. in~\cite{res3d} extended the ResNet architecture with 3D convolutions. Feichtenhofer et al.~\cite{pooling} propose 3D pooling. Sun et al.~\cite{sun3d} decomposed the 3D convolutions into 2D spatial and 1D temporal convolutions. Carreira et al.~\cite{i3d} proposed converting a pre-trained 2D Inception-V1~\cite{googlenet} architecture to 3D by inflating all the filters and pooling kernels with an additional temporal dimension $d$. They achieve this by repeating the weights of 2D filters $d$ times for weight initialization of 3D filters. All these architectures neglect the channel wise information throughout the whole architecture. To the best of our knowledge, our STC block is the first 3D block that integrates channel wise information over 3D networks' layers.

\textbf{Transfer Learning:}  Finetuning or specializing the learned feature representations of a pre-trained network trained on another dataset to a target dataset is commonly referred to as transfer learning. Recently, several works have shown that transferring knowledge within or across modalities (e.g. RGB$\rightarrow$RGB~\cite{hintonrgb} vs. RGB$\rightarrow$Depth~\cite{judyrgbd}, RGB$\rightarrow$Optical-Flow~\cite{judyrgbd,alirgbo}, RGB$\rightarrow$Sound~\cite{l3}, Near-Infrared$\rightarrow$RGB~\cite{rgbir}) is effective, and leads to significant improvements in performance. They typically amount to jointly learning representations in a shared feature space. Our work differs substantially in scope and technical approach. Our goal is to transfer supervision across architectures (i.e. 2D$\rightarrow$3D CNNs), not necessarily limited to transferring information between RGB models only, as our solution can be easily adopted across modalities too.

\section{Proposed Method} \label{sec:method}\emph{}
Our approach with the newly proposed neural block, STC, is to capture different and new descriptor and information in deep CNNs from videos. The spatio-temporal channel correlation block is meant to extract relations between different channels in the different layers of 3D CNNs. The STC block considers these relations in space and time dimensions. In addition, as another major contribution of our work, we show knowledge transfer between cross architectures (i.e. 2D$\rightarrow$3D CNNs), thus avoiding the need to train 3D CNNs from scratch. Details about the transfer learning is given in Section~\ref{subsec:tranferlearning}.

\begin{figure*}[t]
 \centering
 \includegraphics[width=1\columnwidth]{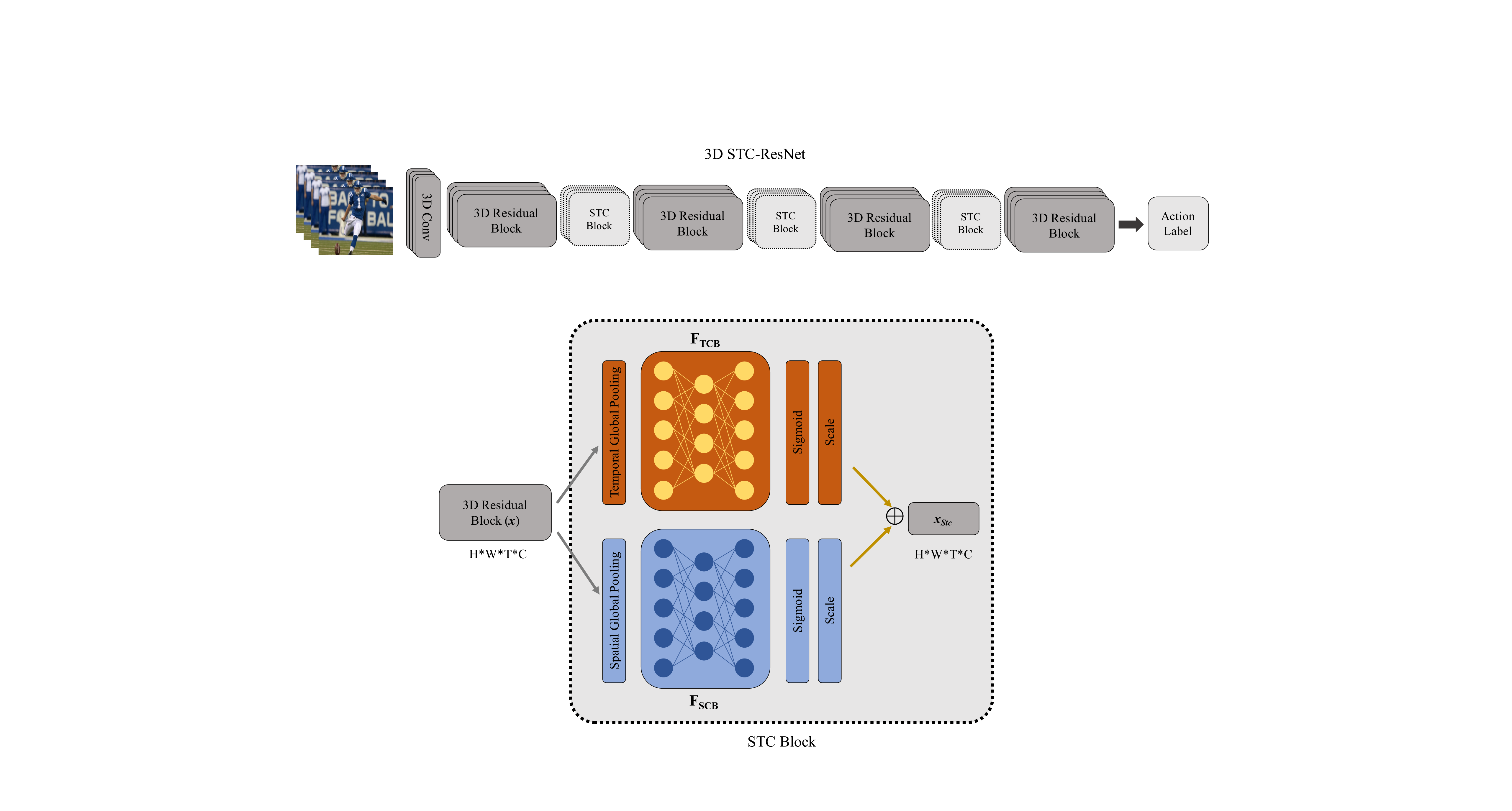}
 \caption{\textbf{STC-ResNet.} Our STC block is applied to the 3D ResNet. The 3D network uses video clips as input. The 3D feature-maps from the clips are densely propagated throughout the network. The STC operates on the different levels  of feature maps in the network to extract spatial and temporal channel relations as new source of information. The output of the network is a video-level prediction.} 
  \label{fig:3dnet}
 \end{figure*}

\subsection{Spatio-Temporal Channel Correlation (STC) Block}
STC is a computational block which can be added to any 3D CNN architecture. Therefore, we have added our STC block to the ResNet and ResNext 3D CNNs introduced by \cite{3DResHara}. After each convolutional block in ResNet and ResNext the STC blocks are inserted to enrich the feature representation. As it was mentioned previously, this new block is exploiting both spatial and temporal information by considering the filters correlation in both spatial and temporal dimension. As input to the STC block, we consider feature maps coming from previous convolution layers. 

\noindent The STC block has a dual path structure which represents different level of concept and information. Each of these paths have different modules; channel or filter  information  embedding and capturing dependencies. To implement we inspired by Squeeze-and-Excitation \cite{senet} method and used global average pooling (spatial and temporal) following with two bottleneck fully connected layers and sigmoid activation. In contrast to \cite{senet}, the STC block has two branches or in other word is dual path; one considering pure channel-wise information and the other takes temporal channel-wise information. Since we are solving video classification, it makes sense to extract more meaningful representations in both spatial and temporal approaches. The STC is capturing channel dependencies information based on this theory. In the following we describe both branches and their contribution to the known 3D architectures like 3D-ResNet \cite{3DResHara}\vspace{0.2cm}.

\textbf{Notation.} The output feature-maps of the 3D convolutions and pooling kernels at the $l^{th}$ layer extracted for an input video is a tensor $X \in \mathbb{R}^{H\times W \times T \times C}$ where $H$, $W$, $T$ and $C$ are the height, width, temporal depth and number of channels of the feature maps, resp. The 3D convolution and pooling kernels are of size (${s\times s \times d}$), where $d$ is the temporal depth and $s$ is the spatial size of the kernels.\vspace{0.2cm} 

\textbf{Temporal Correlation Branch (TCB):}  In this path the feature map will be squeezed by both spatial and temporal dimensions to extract channel descriptors. If we consider $X$ as the input to STC, the output of the first stage, which is a global spatio-temporal pooling is:

\begin{equation}
z_{tcb}=\dfrac{1}{W \times H \times T} \sum_{i}^{W}\sum_{j}^{H}\sum_{t}^{T}{x_{ijt}}
\end{equation}

To obtain the filters non-linear relations, we apply two fully connected layers. The feature dimension is reduced in the first FC layer to $C/r$ (r is reduction ratio) and  is increased again to $C$ by the second FC layer.  Since we used global spatial-temporal pooling over all dimensions of receptive fields, in the next operation, channel-wise information will be extracted. Right after the sigmoid function, the output of the temporal branch ($x_{tcb}$)  will be calculated by rescaling $X$ using $s_{tcb}$  vector. So $s_{tcb}$, output of bottleneck layers, and $x_{tcb}$, the branch output, are calculated as this way: 

\begin{equation}
s_{tcb}=F_{tcb}(z_{tcb},W) = W_{2}(W_{1}z_{tcb})
\end{equation}

\begin{equation}
x_{tcb}=s_{tcb} \cdot X
\end{equation}

\noindent $W$ is the parameter set for the bottleneck layers, including $W_{1} \in \R^{\frac{C}{r} \times C}$ , $W_{2} \in \R^{C \times \frac{C}{r}}$ which are FC layers parameters respectively. $F_{tcb}$ is the symbol of fully-connected functions to calculate the $s_{tcb}$.\vspace{0.2cm}

\textbf{Spatial Correlation Branch (SCB):} The main difference in this branch against the temporal branch is in the aggregation method. The spatial branch shrinks the channel-wise information with respect to the temporal dimension and does global spatial pooling on the input feature map. Therefore this branch is considering the temporal-channel information extraction to enrich the representation in each layer. The calculation of the first operation of the branch comes as following:

\begin{equation}
z_{scb}=\dfrac{1}{W \times H} \sum_{i}^{W}\sum_{j}^{H}{x_{ijT}}
\end{equation}

After the pooling layer, we obtain $z_{scb}$ which is a vector with size of $T\times C$. Afterward, there are the  fully connected layers to extract the temporal based channel relations. In this branch the first FC layer size is $(T\times C)/r$. and second FC size is $C$. Here is the computation description:

\begin{equation}
s_{scb}=F_{scb}(z_{scb},W) = W_{2}(W_{1}z_{scb})
\end{equation}

\begin{equation}
x_{scb}=s_{scb} \cdot X
\end{equation}

which $W_{1} \in \R^{\frac{(T\times C)}{r} \times (T\times C)}$ and $W_{2} \in \R^{C \times \frac{T\times C}{r}}$. By considering both of the branches, the final output of the block ($x_{stc}$) is computed by averaging over the $x_{tcb}$ and $x_{scb}$. In the case of 3D ResNet or ResNext, this output will be added to the residual layer to have the final output of the Convolution (Conv) blocks.
\begin{equation}
x_{stc} = avg(x_{tcb},x_{scb})
\end{equation}

\begin{table}[t]
\caption{\textbf{3D ResNet vs. STC-ResNet and STC-ResNext.} All the proposed architectures incorporate 3D filters and pooling kernels. Each convolution layer shown in the table corresponds the composite sequence BN-ReLU-Conv operations.}
\begin{center}
\resizebox{12cm}{!} {
\begin{tabular}{c|c|c|c|c}
\cline{1-5}
 {Layers}         &  {Output Size}      & 3D-ResNet101 & {3D STC-ResNet101}      & {3D STC-ResNext101}     \\                
\hline \hline
 3D Convolution    & $56\times56\times8$    & \multicolumn{3}{c}{$7\times7\times7$ conv, stride 2}     \\ \hline
 3D Pooling        & $56\times56\times8$      & \multicolumn{3}{c}{$3\times3\times3$ max pool, stride 1} \\ \hline
 
$Res_{1}$   & $28\times28\times8$	 & {$\begin{bmatrix} conv, 1\times1\times1 ,64 \\ conv, 3\times3\times3 ,64 \\ conv, 1\times1\times1 ,256 \end{bmatrix}\times 3$} & {$\begin{bmatrix} conv,1\times1\times1 ,64 \\ conv,3\times3\times3 ,64 \\ conv,1\times1\times1 ,256 \\ fc,[16,256] \end{bmatrix}\times 3$} & {$\begin{bmatrix} conv,1\times1\times1 ,128 \\ conv,3\times3\times3 ,128  \hspace{6mm} C=32 \\ conv,1\times1\times1 ,256 \\ fc,[16,256] \end{bmatrix}\times 3$} \rule{0pt}{0.5ex} \\ \hline

$Res_{2}$    & $14\times14\times4$	 & {$\begin{bmatrix} conv, 1\times1\times1 ,128 \\ conv, 3\times3\times3 ,128 \\ conv, 1\times1\times1 ,512 \end{bmatrix}\times 4$} & {$\begin{bmatrix} conv,1\times1\times1 ,128 \\ conv,3\times3\times3 ,128 \\ conv,1\times1\times1 ,512 \\ fc,[32,512] \end{bmatrix}\times 4$} & {$\begin{bmatrix} conv,1\times1\times1 ,256 \\ conv,3\times3\times3 ,256  \hspace{6mm} C=32 \\ conv,1\times1\times1 ,512 \\ fc,[32,512] \end{bmatrix}\times 4$} \rule{0pt}{0.5ex} \\ \hline

$Res_{3}$   & $7\times7\times2$	 & {$\begin{bmatrix} conv, 1\times1\times1 ,256 \\ conv, 3\times3\times3 ,256 \\ conv, 1\times1\times1 ,1024 \end{bmatrix}\times 23$} & {$\begin{bmatrix} conv,1\times1\times1 ,256 \\ conv,3\times3\times3 ,256 \\ conv,1\times1\times1 ,1024 \\ fc,[64,1024] \end{bmatrix}\times 23$} & {$\begin{bmatrix} conv,1\times1\times1 ,512 \\ conv,3\times3\times3 ,512  \hspace{6mm} C=32 \\ conv,1\times1\times1 ,1024 \\ fc,[64,1024] \end{bmatrix}\times 23$} \rule{0pt}{0.5ex} \\ \hline

$Res_{4}$   & $4\times4\times1$	 & {$\begin{bmatrix} conv, 1\times1\times1 ,512 \\ conv, 3\times3\times3 ,512 \\ conv, 1\times1\times1 ,2048 \end{bmatrix}\times 3$} & {$\begin{bmatrix} conv,1\times1\times1 ,512 \\ conv,3\times3\times3 ,512 \\ conv,1\times1\times1 ,2048 \\ fc,[128,2048] \end{bmatrix}\times 3$} & {$\begin{bmatrix} conv,1\times1\times1 ,512 \\ conv,3\times3\times3 ,512  \hspace{6mm} C=32 \\ conv,1\times1\times1 ,2048 \\ fc,[128,2048] \end{bmatrix}\times 3$} \rule{0pt}{0.5ex} \\ \hline
  
Classification  & $1\times1\times1$			& \multicolumn{3}{c}{$4\times4\times1$ avg pool}  \\  
  \cline{2-5}
  Layer			&  & \multicolumn{3}{c}{400D softmax}     \\ 
\hline

\end{tabular}}
\end{center}
\label{table:arch}\vspace{-0.2cm}
\end{table}

\subsection{Knowledge Transfer} \label{subsec:tranferlearning}
In this section, we describe our method for transferring knowledge between architectures, i.e. pre-trained 2D CNNs to 3D CNNs. Therefore we bypass the need to train the 3D CNNs from scratch with supervision or training with large datasets.

\begin{figure*}[ht]
 \centering
 \includegraphics[width=1\columnwidth]{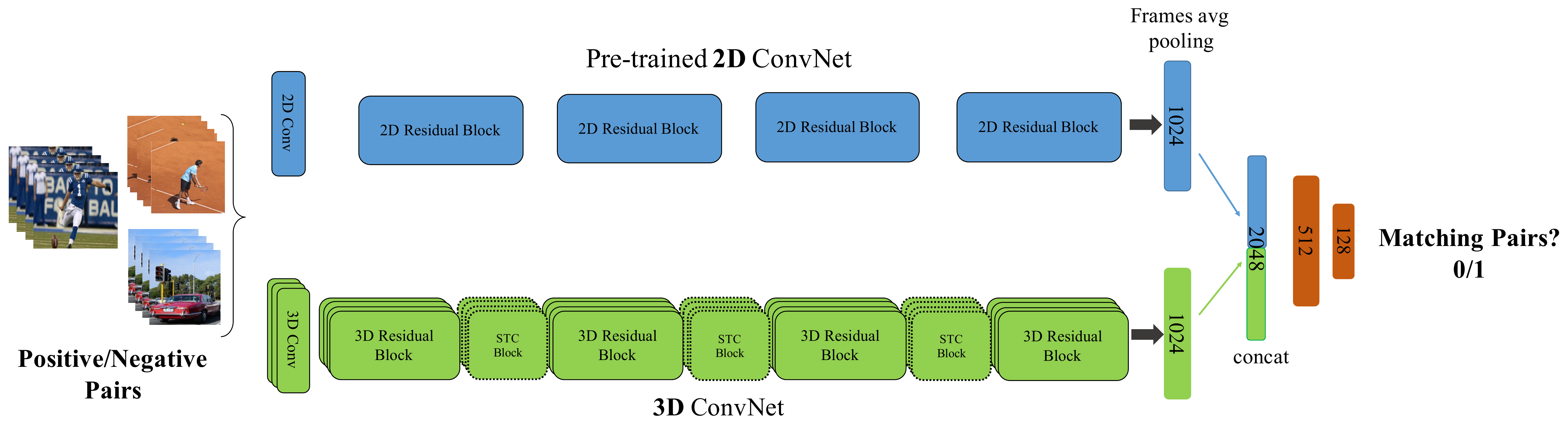}
 \caption{\textbf{Architecture for knowledge transfer from a pre-trained 2D CNN to a 3D CNN.}  The 2D network operates on RGB frames, and the 3D network operates on video clips for the same time stamp. The 2D CNN acts as a teacher for knowledge transfer to the 3D CNN, by teaching the 3D CNN to learn mid-level feature representation by solving an image-video correspondence task. The model parameters of the 2D CNN are frozen, while the task is to effectively learn the model parameters of the 3D CNN only.} 
\label{fig:transfer}\vspace{-0.5cm}
\end{figure*}

Lets assume $\Ib$ is a pre-trained 2D CNN which has learned a rich representation from labeled images dataset, while $\Vb$ being a 3D CNN which is  randomly initialized using~\cite{densenetweighting} and  we want to transfer the knowledge of the representation from $\Ib$ to $\Vb$ for a stable weight initialization. This allows us to avoid training $\Vb$ from scratch, which has million more parameters, and would require heavy computational workload and training time of months~\cite{res3d}. In the current setup, $\Ib$ acts as a teacher for knowledge transfer to the $\Vb$ architecture.

Intuitively, our method uses correspondence between frames and video clips available by the virtue of them appearing together at the same time. Given a pair of $X$ frames and video clip for the same time stamp, the visual information in both frames and video are same. We leverage this for learning mid-level feature representations by an image-video correspondence task between the 2D and 3D CNN architecture, as depicted in Figure~\ref{fig:transfer}. We use 2D ResNet ~\cite{resnet} pre-trained on ImageNet~\cite{imagenet} as $\Ib$, and STC-ResNet network as $\Vb$.  The 2D ResNet CNN has 4 convolution blocks and one fully connected layer at the end, while our 3D architecture has 4 3D-convolution blocks with an STC block and we add a fully-connected layer after the last block.  We concatenate the last $fc$ layers of both architectures, and connect them  with the \textbf{2048-dimensional fc} layer which is in turn connected to two fully connected layers with 512 and 128 sizes (fc1 , fc2) and to the final binary classifier layer.  We use a binary matching classifier: given $X$ frames and a video clip, decide whether the pairs belong to each other or not. For a given pair, $X$ frames are fed sequentially into the network $\Ib$ and we average the last 2D fc features over the $X$ frames, resulting into \textbf{1024-D} feature representation. In parallel the video clip is fed to the network $\Vb$, and we extract the 3D fc features (1024-D), and concatenate them, which is then passed to the fully connected layers for classification. For training, we use a binary classification loss.

During the training, the model parameters of $\Ib$ are frozen, while the task is to effectively learn the model parameters of $\Vb$ without any additional supervision than correspondences between frames and video. The pairs belonging to the same time stamp from the same video are positive pairs, while the pairs coming from two different videos by randomly sampling $X$ frames and video clips from two different videos is a negative pair. Note that, during back-propagation, only the model parameters for $\Vb$ are updated, i.e., transferring the knowledge from $\Ib$ to $\Vb$. In our experiments we show that a stable weight initialization of $\Vb$ is achieved, and when fine-tuned on the target dataset, allows the model to adapt quickly to the target dataset, thus avoiding training the model from scratch with improved performance. We also show that by using our proposed knowledge transfer method, 3D CNNs can be trained directly on small datasets like UCF101 and achieve a better performance than training from scratch.

Since our transfer learning is unsupervised and there is no need of video label, we have applied on a collection of unlabeled videos. Further, our experiments in Section~\ref{sec:experiments} demonstrate that our proposed transfer learning of STC-ResNext outperforms the generic 3D CNNs by a significant margin which was trained on large video dataset, Sports-1M~\cite{sport1m} and finetuned on the target datasets, HMDB51/UCF101.


\section{Experiments} \label{sec:experiments}
In this section, we first introduce the datasets and implementation details of our proposed approach. Afterwards, we demonstrate an extensive study on the architecture of the proposed STC-ResNet and STC-ResNext which are 3D CNNs, and then the configurations for input data. Following, we evaluate and compare our proposed methods with the baselines and other state-of-the-art methods. Finally, we compare our transfer learning: $2D\rightarrow3D$ CNN performance with generic state-of-the-art 3D CNN methods.  For the ablation study of architecture and configurations of input data,  we  report  the  accuracy  of split 1 on UCF101.
\subsection{HMDB51, UCF101, and Kinetics Datasets}

We evaluate our proposed method on three challenging video datasets with human actions, namely HMDB51~\cite{hmdb51}, UCF101~\cite{ucf101}, and Kinetics~\cite{i3d}. Table~\ref{table:databases} shows the details of the datasets. For all of these datasets, we use the standard training/testing splits and protocols provided as the original evaluation scheme. For HMDB51 and UCF101, we report the average accuracy over the three splits and for Kinetics, we report the performance on the validation and test set.

\textbf{\textbf{Kinetics:}}
Kinetics is a new challenging human action recognition dataset introduced by~\cite{i3d}, which contains 400 action classes. There are two versions of this dataset: untrimmed and trimmed. The untrimmed videos contain the whole video in which the activity is included in a short period of it. However, the trimmed videos contain the activity part only. We evaluate our models on the trimmed version. We use the whole training videos for training our models from scratch. Our results for both the STC-ResNet101 and STC-ResNext101 models are reported in the Table~\ref{table:kinetics_results}.

\textbf{\textbf{UCF101:}}
For evaluating our STC-Nets architectures, we first trained them on the Kinetics dataset, and then fine-tuned them on the UCF101. Furthermore, we also evaluate our models by training them from scratch on UCF101 using randomly initialized weights to be able to investigate the effect of pre-training on a huge dataset, such as Kinetics. 

\textbf{\textbf{HMDB51:}}
Same as UCF101 evaluation we fine-tune the models on HMDB51, which were pre-trained from scratch on Kinetics. Similarly, also we evaluate our models by training them from scratch on HMDB51 using randomly initialized weights.

\begin{table}[hbt]
\caption{Details of the datasets used for evaluation. The `Clips' shows the total number of short video clips extracted from the `Videos' available in the dataset.}
{\small
\tabcolsep=0.3cm
\begin{center}
\begin{tabular}{ l |   c     c     c}
\hline
Data-set &  \# Clips &\# Videos &\# Classes\\
\hline
\hline
HMDB51~\cite{hmdb51}  		& 6,766  &	3,312 & 51\\
UCF101~\cite{ucf101}  		& 13,320 &	2,500& 101\\
Kinetics~\cite{i3d}   		& 306,245&	306,245& 400\\
\hline
\end{tabular}
\end{center}}
\label{table:databases}\vspace{-0.5cm} 
\end{table}


\subsection{Implementation Details}
We use the PyTorch framework for 3D CNNs implementation and all the networks are trained on 8 Tesla P100 NVIDIA GPUs. Here, we describe the implementation details of our two schemes, 3D CNNs and  knowledge transfer from 2D to 3D CNNs for stable weight initialization.

\noindent
\paragraph{\textbf{Training:}} \emph{}

\noindent
\textbf{$-$ STC-Nets.} We train our STC-Nets (STC-ResNet/ResNext) from scratch on Kinetics. Our STC-Net operates on a stack of 16/32 RGB frames. We resize the video to 122px when smaller, and then randomly apply 5 crops (and their horizontal flips) of size $112\times112$.  For the network weight initialization, we adopt the same technique proposed in~\cite{densenetweighting}. For the network training, we use SGD, Nesterov momentum of 0.9, weight decay of $10^{-4}$ and batch size of 128.  The initial learning rate is set to 0.1, and reduced by a factor of 10x manually when the validation loss is saturated. The maximum number of epochs for the whole Kinetics dataset is set to 200. Batch normalization also has been applied. Also the reduction parameter in STC blocks, $r$ is set to 4. 
\noindent
\paragraph{\textbf{Testing:}}  For video prediction, we decompose each video into non-overlapping clips of 16/32 frames. The STC-Net is applied over the video clips by taking a $112\times112$ center-crop, and finally we average the predictions over all of the clips to make a video-level prediction.

\noindent
\textbf{$-$ Knowledge Transfer: $2D\rightarrow3D$ CNNs.} We employ 2D ResNet and ResNext architectures, pre-trained on ImageNet~\cite{imagenet},while the 3D CNN is our STC-ResNet and 3D-ResNet network. To the 2D CNN, 16 RGB frames are fed as input. The input RGB images are randomly cropped to the size $112 \times 112$, and then mean-subtracted for the network training. To supervise transfer to the STC-ResNet, we replace the previous classification layer of 2D CNN  with a $2$-way softmax layer to distinguish between positive and negative pairs.  We use stochastic gradient descent (SGD) with mini-batch size of 32 with a fixed weight decay of $10^{-4}$ and Nesterov momentum of 0.9. For network training, we start with  learning rate set to 0.1 and  decrease it by a factor of 10 every 30 epochs. The maximum number of epochs is set to 150.


\subsection{Ablation Study on Architecture Design}
To evaluate our STC block on 3D CNNs model, we conducted an architecture study and evaluated different configurations.
For this work, we mainly focused on 3D versions for ResNet and ResNext with different input size and depth. Our choice is based on the recently presented nice performance of these networks in video classification \cite{3DResHara}.

\paragraph{\textbf{Model Depth:}}
We first analyze the impact of the architecture depth with 3D-ResNet and 3D-ResNext and we have done a series of evaluations on the network-size, and temporal-depth of input data to the network with the new STC block. For the architecture study, the model weights were initialized using~\cite{densenetweighting}.

We employ three different sizes of 3D STC-ResNet; 18, 50, 101 with STC blocks. Evaluations results of these 3D STC-ResNet models are reported in the Table~\ref{table:modelDepth}. As it can be observed, by adding small overhead of STC blocks, STC-Nets can achieve reasonable performance even in smaller version of ResNet, since our STC-ResNet50 is comparable with regular ResNet101.    

\begin{table}[hbt]
\caption{Evaluation results of 3D STC-ResNet model with network sizes of 18, 50, and 101 on UCF101 split 1. All models were trained from scratch.}
{\small
\tabcolsep=0.3cm
\begin{center}
\begin{tabular}{ c |   c}
\hline
Model Depth&   Accuracy~\%\\
\hline
\hline
3D-ResNet 101  		& 46.7\\
\hline
STC-ResNet 18  		& 42.8\\
STC-ResNet 50  		& 46.2\\
STC-ResNet 101  		& \textbf{47.9}\\
\hline
\end{tabular}
\end{center}}
\label{table:modelDepth}
\end{table}

Temporal depth of series of input frames plays a key role in activity recognition tasks. Therefore, we have reported the performance of our 3D STC-ResNet with  different configuration of temporal depths in the Table~\ref{table:temporalDepth}. Our evaluation shows the fact that longer clips as input will yield better performance, and similarly also presented the same fact in \cite{3DResHara,i3d}.

\begin{table}[h]
\caption{Evaluation results of 3D STC-ResNet model with temporal depths of 16 and 32 on UCF101 split 1. All models were trained from scratch.}
{\small
\tabcolsep=0.3cm
\begin{center}
\begin{tabular}{ c |   c}
\hline
Temporal Depth&   Accuracy~\% \\
\hline
\hline
16  		& 45.6\\
32  		& \textbf{47.9}\\
\hline
\end{tabular}
\end{center}}
\label{table:temporalDepth}\vspace{-1cm} 
\end{table}

\paragraph{\textbf{TCB vs SCB:}} We also have studied the impact of the TCB and SCB branches in our STC-Nets. Since each of them consider different concepts in the branch, we evaluated the performance in three settings: SCB only, TCB only, and SCB-TCB combination (STC). In the Table~\ref{TCB vs SCB}, the importance of the channel correlation branches is shown. As it is shown, incorporating both branches to capture different descriptors is performing better than single cases.

\begin{table}[hbt]
\caption{Performance comparison using different channel correlation blocks (TCB vs SCB).}
{\small
\tabcolsep=0.3cm
\begin{center}
\begin{tabular}{ c |   c}
\hline
Channel Correlation Branch &   Accuracy~\% \\
\hline
\hline
SCB 		& 46.1\\
TCB 		& 47.2\\
TCB \texttt{+} SCB    & \textbf{47.9} \\
\hline
\end{tabular}
\end{center}}
\label{TCB vs SCB}\vspace{-0.5cm} 
\end{table}

\paragraph{\textbf{Frame Sampling Rate:}}
Finding right configuration of input-frames which are fed to the CNNs for capturing the appearance and temporal information plays a very critical role in temporal CNNs. For this reason, we investigated frame sampling rate for the input stream. 
The STC-ResNet101 has been used for ablation study on frame sampling rate for training and testing phase. We evaluate the model by varying the temporal stride of the input frames in the following set \{1, 2 ,4, 16\}. Table~\ref{table:frameSamplingRate} presents the accuracy of STC-ResNet101 trained on inputs with different sampling rates. The best results are obtained with sampling rate of 2, which we also used for other 3D CNNs in the rest of the experiments: STC-ResNet101 and 3D-ResNet101.

\begin{table}[hbt]
\caption{Evaluation results of different frame sampling rates for STC-ResNet101 model. Trained and tested on UCF101 split 1.}
{\small
\tabcolsep=0.3cm
\begin{center}
\begin{tabular}{ l ||    c c c c }
\hline
Input Stride  & 1  &2 &4 &16\\
\hline
Accuracy~\%  	& 44.6\% &\textbf{47.9}\% &46.8\% &40.3\%\\
\hline
\end{tabular}
\end{center}}
\label{table:frameSamplingRate}\vspace{-1cm} 
\end{table}

\subsection{Knowledge Transfer} 
To apply our proposed supervision transfer, we have tested 2D ResNet and ResNext as basic pre-trained model on ImageNet, while 3D-ResNet and our STC-ResNet with randomly initialized using~\cite{densenetweighting}, as target 3D CNNs.  We show that, a stable weight initialization via transfer learning is possible for 3D CNN architecture, which can be used as a good starting model for training on small datasets like UCF101.

Here, we explain the training phase for STC-ResNet (see Fig.~\ref{fig:transfer}) case which is similar to the other networks.  To train, we have negative and positive video clip pairs to feed to the networks. Given a pair of 16 frames, video clips for the same time stamp will go through the 2D ResNet and STC-ResNet. For the 2D network whose model weights are frozen, we do average pooling on the last layer with size of 1024. So, pooled frame features from 2D network are concatenated with clip feature from 3D network ($1024+1024$), and passed to 2 fully connected layers afterward. The fully connected layer sizes are 512, 128. The binary classifier distinguishes between correspondence of negative and positive clip pairs. The STC-ResNet network is trained via back-propagation through the network, and the 3D kernels are learned.

\begin{table}[ht]
\caption{Transfer learning results for 3D CNNs by 2D CNNs. Both of the results are on UCF101 split1. First column shows the performance of transfered network finetuned directly on UCF101. The second column is finetuned transfered network first on the half of Kinetics dataset and then on UCF101.}
{\small
\tabcolsep=0.3cm
\begin{center}
\begin{tabular}{ l | c | c  }
\hline
\textbf{3D CNNs} &  \textbf{Transfer} & \textbf{FT-Transfer} \\
\hline
3D-ResNet 			& 82.1 & 84.6\\ \hline
3D-ResNext			& 83.1 & 86.9\\ \hline
\hline
STC-ResNet					& 83.2 & 86.5\\ \hline
STC-ResNext					& \textbf{84.7} & \textbf{88.2}\\ \hline
\end{tabular}
\end{center}}
\label{table:Trasnsfer_result}
\end{table}


Another important aspect of proposing this transfer learning for 3D CNNs is finding a cheaper way to train 3D CNNs when availability of large datasets is at scarce. After pre-training our 3D CNNs by  described transfer learning, we can use a fraction of a big dataset (e.g. Kinetics) to train the model and still achieve a good performance in fine-tuning on UCF101. In other words, this knowledge transfer reduces the need for more labeled data and very large datasets. To perform transfer learning, we use approx. 500K unlabeled videos from YouTube8m dataset~\cite{youtube8m}. Since the transfer learning pipeline for 3D CNNs have been tested with three different deep architectures (3D-ResNet, 3D-ResNext and STC-Nets), we clearly show the generalization capacity of our method in deep architectures, which can be easily adopted for other deep networks and tasks which use the similar architectures. Table~\ref{table:Trasnsfer_result} shows the results, we can observe that via transfer learning we achieve better performance in comparison to training the network from scratch.

\begin{table}[hb] 
\caption{Comparison results of our models with other state-of-the-art methods on Kinetics dataset. * denotes the pre-trained version of C3D on the Sports-1M.}
{\small
\tabcolsep=0.3cm
\begin{center}
\begin{tabular}{ l |   c |  c }
\hline
\textbf{Method} &  \textbf{Top1-Val} &  \textbf{Top5-Val} \\
\hline
DenseNet3D 						& 59.5 & -\\ \hline
Inception3D 						& 58.9 &-\\ \hline
C3D*~\cite{3DResHara}			& 55.6 	&-\\ \hline
3D ResNet101~\cite{3DResHara}		& 62.8 	& 83.9\\ \hline
3D ResNext101~\cite{3DResHara}		& 65.1 	& 85.7\\ \hline
RGB-I3D~\cite{i3d}		& 68.4 	& 88\\ \hline
\hline\hline
\textbf{STC-ResNet101}					& 64.1 	&85.2\\ \hline
\textbf{STC-ResNext101}					& 66.2 	&86.5\\ \hline
\textbf{STC-ResNext101 32f}					& \textbf{68.7} 	& \textbf{88.5}\\ \hline
\end{tabular}
\end{center}}
\label{table:kinetics_results}\vspace{-0.5cm}
\end{table}

\subsection{Comparison with the state-of-the-art}
Finally, after exploring and studying on STC-Net architectures and the configuration of input-data and architecture, we compare our STC-ResNet and STC-ResNext with the state-of-the-art methods by pre-training on Kinetics and finetuning on all three splits of the UCF101 and HMDB51 datasets. For the UCF101 and HMDB51, we report the average accuracy over all three splits. The results for supervision transfer technique experiments were reported in the previous part of experiments.

Table~\ref{table:kinetics_results} shows the result on Kinetics dataset for STC-Nets compared with state-of-the-art methods. C3D~\cite{c3d} employs batch normalization after each convolutional and fully connected layers (C3D w/ BN), and RGB-I3D which is without pretraining on the ImageNet~(RGB-I3D w/o ImageNet)~\cite{i3d}. The STC-ResNet achieves higher accuracies than 3D ResNet101, Sports-1M pre-trained C3D and C3D w/ BN which is trained from scratch. However, RGB-I3D achieved competitive performance which might be the result of usage of longer video clips than ours (64 vs. 32), Although we trained our own version of Inception3D same as I3D~\cite{i3d},but achieved  different results due to difference experimental setup. As mentioned earlier, due to high memory usage of 3D models we had to limit our model space search and it was not possible to checkout the longer input video clips. Moreover,~\cite{i3d} used larger number of mini-batches by engaging a large number of 64 GPUs that they have used, which plays a vital role in batch normalization and consequently training procedure.

\begin{table}[t]
\caption{Accuracy (\%) performance comparison of STC-Nets (STC-ResNet/ResNext) with state-of-the-art methods over all three splits of UCF101 and HMDB51.}
{\small
\tabcolsep=0.3cm
\begin{center}
\begin{tabular}{ l |   c |  c   }
\hline
\textbf{Method} &  \textbf{UCF101} & \textbf{HMDB51}\\
\hline
DT+MVSM~\cite{dtmvsv}				& 83.5 & 55.9	\\ \hline
iDT+FV~\cite{idt}					& 85.9 & 57.2	\\ \hline
C3D~\cite{c3d}						& 82.3 & 56.8	\\ \hline
Conv Fusion~\cite{pooling}					& 82.6 & 56.8	\\ \hline
Two Stream~\cite{twostream}		& 88.6 & $-$	\\ \hline
TDD+FV~\cite{tdd}					& 90.3 &	63.2	\\ \hline
TSN-RGB~\cite{tsn}						& 85.7 & $-$	\\ \hline
P3D~\cite{p3d}	& 88.6 & $-$	\\ \hline
\hline
Inception3D	& 87.2 & 56.9	\\ \hline
I3D \cite{i3d}	& 95.6 & \textbf{74.8}	\\ \hline
3D ResNet 101	& 88.9 & 61.7	\\ \hline
3D ResNet 101-Transfered Knowledge & 91.3 & 64.2	\\ \hline
3D ResNext 101	& 90.7 & 63.8	\\ \hline

\hline
\textbf{STC-ResNet 101 }				& {90.1}		&  {62.6}	\\ \hline
\textbf{STC-ResNext 101 }	& {92.3}		&  {65.4}	\\ \hline
\textbf{STC-ResNet 101-Transfered Knowledge }	& {92.6}		&  {66.1}	\\ \hline
\hline
\textbf{STC-ResNet 101 (64 frames) }	& {93.7}		&  {66.8}	\\ \hline
\textbf{STC-ResNext 101 (32 frames)}	& \textbf{95.8}		&  72.6	\\ \hline
\end{tabular}
\end{center}}
\label{table:state_comparison_UCFHMDB}
\end{table}

Table~\ref{table:state_comparison_UCFHMDB} shows the results on UCF101 and HMDB51 datasets for comparison of STC-Nets with other RGB based action recognition methods. Our STC-ResNext101 models outperform the 3D-ResNet~\cite{res3d}, Inception3D and C3D~\cite{c3d} on both UCF101 and HMDB51 by 95.8\% and 72.6\%  respectively. As mentioned before we trained Inception3D, a similar architecture to the I3D~\cite{i3d} (without using ImageNet) on Kinetics and fine-tuned it on UCF101 and HMDB51 to be able to have a more fair comparison. As shown in the Table~\ref{table:state_comparison_UCFHMDB}, STC-ResNext performs better than 3D-ResNext by almost 2\% on UCF101. Furthermore, STC-ResNext and STC-ResNet achieve the best performance among the methods using only RGB input on UCF101 and HMDB51. Moreover it should be noted that, the reported result of RGB-I3D~\cite{i3d} pre-trained on ImageNet and Kinetics by Carreira et al.~\cite{i3d} has close and comparable result to ours on both UCF101 and HMDB51, this might be due to difference in usage of higher input resolution and larger mini-batch sizes by using 64 GPUs . Furthermore, we note that the state-of-the-art CNNs~\cite{i3d,tsn} use  expensive optical-flow maps in addition to RGB input-frames, as in I3D which obtains a performance of 98\% on UCF101 and 80\% on HMDB51. Because of such a high computation needs, we are not able to run the similar experiments, but as it can be concluded from Table~\ref{table:state_comparison_UCFHMDB}, our best RGB model has superior performance than the other RGB based models.

Note that, in our work we have not used dense optical-flow maps, and still achieving comparable performance to the state-of-the-art methods~\cite{tsn}. This shows the effectiveness of our STC-Nets to exploit temporal information and spatio-temporal channel correlation in deep CNNs for video clips. This calls for efficient methods like ours instead of computing the expensive optical-flow information (beforehand) which is very computationally demanding, and also difficult to obtain for large scale datasets.

\section{Conclusion} \label{sec:conclusion}

In this work, we introduce a new `Spatio-Temporal Channel Correlation'~(STC) block that models correlation between the channels of a 3D CNN. We clearly show the benefit of exploiting spatio-temporal channel correlations features using STC block. We equipped 3D-ResNet and 3D-ResNext with our STC block and improved the accuracies by 2-3\% on Kinetics dataset. We name our architecture as STC-Nets. Our STC blocks are added as a residual unit to other parts of networks and learned in an end-to-end learning. The STC feature-maps model the feature interaction in a more expressive and efficient way without an undesired loss of information throughout the network. Our STC-Nets are evaluated on three challenging action recognition datasets, namely HMDB51, UCF101, and Kinetics. STC-Nets architecture achieves state-of-the-art performance on HMDB51, UCF101 and comparable results on Kinetics , in comparison to other temporal deep neural network models. Even though, our STC block has the potential to generalize to any other 3D architecture too. Further, we show the benefit of transfer learning between cross architectures, specifically supervision transfer from 2D to 3D CNNs. This provides a valuable and stable weight initialization for 3D CNNs instead of training it from scratch and this also avoids the computational costs. However, our transfer learning approach is not limited to transfer supervision between RGB models only, as our solution can be easily adopted across modalities too.

\bibliographystyle{splncs}
\bibliography{egbib}
\end{document}